\documentclass[10pt]{article}

\usepackage{amsmath}
\usepackage{amsthm}
\usepackage{graphicx}
\usepackage{enumitem}
\usepackage{algorithm}
\usepackage{cite}
\usepackage{color}
\usepackage{bm}
\usepackage[noend]{algpseudocode}
\usepackage{titlesec}
\usepackage{fancyhdr}
\usepackage{lastpage}
\usepackage{fullpage}
\usepackage{lipsum}
\usepackage{mathtools}

\usepackage{epstopdf}
\DeclareGraphicsExtensions{.eps,.pdf,.png,.jpg}
\usepackage{pgf}
\usepackage{import}
\graphicspath{{gfx/}}

\usepackage[T1]{fontenc}
\usepackage{textcomp}
\usepackage[charter]{mathdesign}
\usepackage{oldgerm}

\let\pgfimageWithoutPath\pgfimage 
\renewcommand{\pgfimage}[2][]{\pgfimageWithoutPath[#1]{gfx/#2}}

\titleformat*{\section}{\large\bfseries}
\titleformat*{\subsection}{\normalsize\bfseries}

\newtheorem{defn}{Definition}

\theoremstyle{remark}

\newcommand{\NN}{\mathbb{N}}

\newcommand{\RR}{\mathbb{R}}

\newcommand{\EE}{\mathbb{E}}

\newcommand{\Gg}{\mathcal{G}}

\newcommand{\Jj}{\mathcal{J}}

\newcommand{\al}{\alpha}
\newcommand{\la}{\lambda}

\renewcommand{\phi}{\varphi}

\DeclareMathOperator{\diag}{diag}

\DeclareMathOperator{\prox}{prox}

\newcommand{\norm}[1]{ \left \| #1 \right \|}

\newcommand{\Id}{\mathbf{I}}

\newcommand{\uargmin}[1]{\underset{#1}{\argmin}\;}

\newcommand{\ie}{\emph{i.e.}, }

\newcommand{\tr}{{\tiny{\mathrm{T}}}}

\newcommand{\bs}{\boldsymbol}

\newcommand{\bal}{\bs \al}

\newcommand{\hi}{\mathrm{hi}}
\newcommand{\lo}{\mathrm{lo}}

\def\defn{\,\triangleq\,}
\newcommand{\iter}[1]{\!\!\;#1}
\newcommand{\closeiter}[1]{\!\!#1}

\def\dbf{{\mathbf{d}}}

\def\xbf{{\mathbf{x}}}
\def\ybf{{\mathbf{y}}}

\def\ebf{{\mathbf{e}}}

\def\bbf{{\mathbf{b}}}
\def\cbf{{\mathbf{c}}}

\def\alphabm{\bm{\alpha}}
\def\phibm{\bm{\phi}}

\def\xbfhat{{\widehat{\xbf}}}
\def\alphabmhat{{\widehat{\alphabm}}}

\def\Phibf{{\mathbf{H}}}
\def\Dbf{{\mathbf{D}}}

\def\Abf{{\mathbf{A}}}

\def\Lbf{{\mathbf{L}}}
\def\Cbf{{\mathbf{C}}}
\def\Sbf{{\mathbf{S}}}

\def\Rcal{{\mathcal{R}}}
\def\Ccal{{\mathcal{C}}}
\def\Lcal{{\mathcal{L}}}

\def\R{\mathbb{R}}

\def\Dfrak{\mathfrak{D}}

\def\argmin{\mathop{\mathrm{arg\,min}}}
\def\vert{\mathrm{vc}}

\setlist[itemize,1]{leftmargin=\dimexpr 26pt-0.5cm}
\renewcommand{\algorithmicrequire}{\textbf{Input:}}

\begin{document}


\title{Online Convolutional Dictionary Learning for\\ Multimodal Imaging}


\author{K\'{e}vin~Degraux%
\thanks{K.~Degraux (email:~kevin.degraux@uclouvain.be) is with ISPGroup/ICTEAM, FNRS, Universit{\'e}
  catholique de Louvain, 1348, Louvain-la-Neuve, Belgium. This work
was completed while he was with Mitsubishi Electric Research Laboratories (MERL).}
\hspace{0.05em},
Ulugbek~S.~Kamilov%
\thanks{U.~S.~Kamilov (email:~kamilov@merl.com), P.~T.~Boufounos (email:~petrosb@merl.com), and D.~Liu (email:~liudh@merl.com)
are with Mitsubishi Electric Research Laboratories (MERL), 201 Broadway, Cambridge,
MA 02139, USA.}
\hspace{0.05em}, Petros~T.~Boufounos$^\dagger$, and Dehong~Liu$^\dagger$}    

\maketitle


\begin{abstract}
Computational imaging methods that can exploit multiple modalities have the potential to enhance the capabilities of traditional sensing systems. In this paper, we propose a new method that reconstructs multimodal images from their linear measurements by exploiting redundancies across different modalities.
Our method combines a convolutional group-sparse representation of images with total variation (TV) regularization for high-quality multimodal imaging. We develop an online algorithm that enables the unsupervised learning of convolutional dictionaries on large-scale datasets that are typical in such applications. We illustrate the benefit of our approach in the context of joint intensity-depth imaging.
\end{abstract}


\section{Introduction}
\label{sec:intro}

Multimodal imaging systems acquire several measurements of an object
using multiple distinct sensing modalities. Often, the data acquired
from the sensors is jointly processed to improve the imaging quality
in one or more of the acquired modalities. Such imaging methods have
the potential to enable new capabilities in traditional sensing
systems, providing complementary sources of information about the
object. Some of the most common applications of multimodal imaging
include remote sensing~\cite{Ma.etal2014}, biomedical
imaging~\cite{Fatakdawala.etal2013}, and high-resolution depth
sensing~\cite{Diebel.Thrun2005}.

We consider a joint imaging inverse problem with multiple noisy linear measurements
\begin{align}
    \ybf_\ell  = \Phibf_\ell \xbf_\ell + \ebf_\ell,
\end{align}
where for each modality $\ell \in [1, \dots, L]$, $\ybf_\ell \in\R^{M_\ell}$ denotes the corresponding measurement vector, $\xbf_\ell \in \R^N$ denotes the unknown image, $\Phibf_\ell \in \R^{{M_\ell}\times N}$ denotes the sensing matrix, and $\ebf_\ell \in \R^{M_\ell}$ denotes the noise in the measurements.
The images $\{\xbf_\ell\}_{\ell \in [1 \dots L]}$ correspond to the same physical object viewed from different modalities. For example, each $\xbf_\ell$ may represent a different color channel, spectral band, or a type of sensor. 
For simplicity, we assume that the desired dimension of the images is the same across all modalities and that acquisition devices are perfectly registered. 
The key insight used in our paper is that information about a single modality exists, in  some form, in other modalities. This information
can be exploited to improve the quality of multimodal imaging, as long
as it can be extracted from the measurements.

\subsection{Main Contributions}

\begin{figure}[t]
\centering
\vspace{-2mm}
\includegraphics[width = 0.47\textwidth]{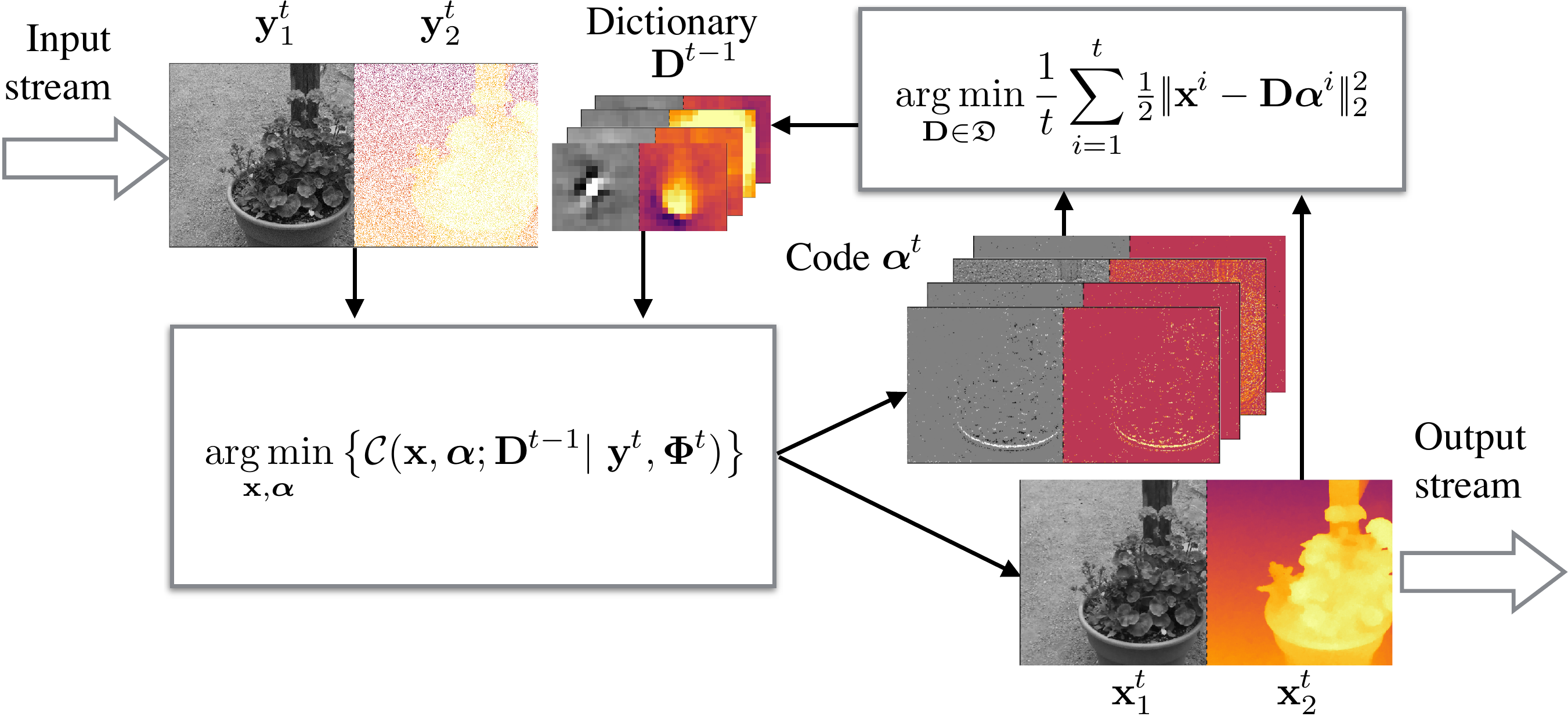}\vspace{-4mm}
\caption{Illustration of the proposed multimodal imaging method.}
\vspace{-3mm}
\label{fig:illustration}
\end{figure}

In this work, we propose a novel approach based on  jointly sparse representation of multimodal images. Specifically, we are interested in learning data-adaptive convolutional dictionaries for both reconstructing and representing the signals given their linear measurements. The main benefit of a convolutional approach is that it is translation invariant and leads to a sparse representation over the entire image. This, however, comes with the increase in the computational cost, which we address by developing a new online convolutional dictionary learning method suitable for working with large-scale datasets. Our key contributions are summarized as follows:

\begin{itemize}

\item  We provide a new formulation for multimodal computational
  imaging, incorporating a convolutional joint sparsity prior and a total variation (TV) regularizer. In this formulation, the high resolution images are determined by solving an optimization problem, where the regularizer exploits the redundancies across different modalities.

\item We develop an online convolutional dictionary learning algorithm,
  illustrated in Figure~\ref{fig:illustration}. By accommodating an
  additional TV regularizer in the cost, the algorithm is able to learn the convolutional dictionary in an unsupervised fashion, directly from the noisy measurements. We validate our approach for joint intensity-depth imaging.

\end{itemize}

\vspace{-4mm}
\subsection{Related Work}

Starting from early work by Olshausen and
Field~\cite{Olshausen.Field1996, Olshausen1997}, dictionary learning
has become a standard tool for various tasks in image
processing~\cite{Aharon.etal2006, Mairal.etal2009, Yang.etal2012,
  Wei2014, Tillmann.etal2016}. Our approach builds upon two prior
lines of research, one on convolutional sparse
representations~\cite{Zeiler.etal2010, Yang.etal2010, Wohlberg2016}
and one on online dictionary learning~\cite{Mairal2010, Mensch2016a,
  Mensch2016}. Since our method relies on TV regularization, it is
also related to TV-based imaging algorithms~\cite{Rudin1992, Beck2009,
  Afonso2010, Kamilov2017}. Specifically, our method is based on the
popular fast iterative shrinkage/thresholding algorithm (FISTA) for
reconstructing images from measurements. Our method is validated on
the problem of joint intensity--depth imaging, also considered
in~\cite{Kopf.etal2007, He.etal2013, Tosic2009, Tosic2014,
  Castorena2016a, Kamilov2016}. In particular,~\cite{Tosic2009,
  Tosic2014} use traditional sparse coding for combining depth and
intensity, and~\cite{Liu2016} uses convolutional dictionaries for
representing multiple modalities. Our approach extends earlier work by
performing multi-modal image reconstruction with convolutional
dictionaries and developing a dedicated online learning algorithm for
large-scale settings.


\section{Proposed Method}
\label{sec:method}
\subsection{Problem Formulation}
\label{sec:problem}
The underlying assumption in our approach is that a jointly sparse
convolutional model can accurately approximate the images
$\{\xbf_\ell\}$ as
\begin{equation}
 \xbf_\ell \approx \Dbf_\ell \alphabm_\ell \defn \sum_{k = 1}^K \dbf_{\ell k} \ast \alphabm_{\ell k},
\end{equation}
where $\{\dbf_{\ell k}\}$ is the set of $LK$ convolutional filters in $\RR^P$, $\ast$~denotes convolution, and $\{\alphabm_{\ell k}\}$ is the set of coefficient maps in $\RR^{\hat{N}}$. Note that $\Dbf_\ell$ and $\alphabm_\ell$ denote the concatenation of all $K$ dictionaries and coefficient maps, respectively. Given the complete dictionary $\Dbf = (\Dbf_1, \dots, \Dbf_L)$, we can define our imaging problem as the following joint optimization
\begin{equation}
\label{Eq:SparseCoding}
 (\xbfhat, \alphabmhat) = \argmin_{\xbf, \alphabm} \left\{\Ccal(\xbf, \alphabm; \Dbf \,|\, \ybf,\Phibf)\right\},
\end{equation}
where the cost function $\Ccal$ is given by
\begin{align}
\label{Eq:CostFunction}
 \Ccal(\xbf, \alphabm; \Dbf \,|\, \ybf,\Phibf) 
&\defn  \tfrac{1}{2}\|\ybf - \Phibf \xbf\|_{2}^2 + \tfrac{\rho}{2}\|\xbf - \Dbf\alphabm\|_{2}^2 \\
&\quad\quad+ \lambda \|\alphabm\|_{2,1} + \tau \Rcal(\xbf), \nonumber
\end{align}
with 
$\ybf \defn \vert(\ybf_1, \dots, \ybf_L)$, $\xbf \defn \vert(\xbf_1,
\dots, \xbf_L)$, and $\alphabm \defn \vert(\alphabm_1, \dots,
\alphabm_L)$, denoting the vertical concatenation (vc) of
corresponding signals and $\Phibf \defn \diag(\Phibf_1, \dots,
\Phibf_L)$ denoting the block diagonal sensing matrix. The first
quadratic term in~\eqref{Eq:CostFunction} measures the data-fidelity,
while the second controls the approximation quality of the
dictionaries. The first regularization term
\begin{equation}
\|\alphabm\|_{2, 1} \defn \sum_{k = 1}^K \sum_{n = 1}^{\hat{N}} \norm{ \alphabm_{\cdot kn} }_2 
\end{equation}
imposes group- or joint-sparsity of coefficients across $L$
modalities. Here, $\alphabm_{\cdot kn} \in \RR^L$ denotes the vector
formed by the aligned entries of the coefficient maps associated with
kernel $k$ for every modality $\ell$.  Specifically, this regularizer
promotes the co-occurrence of image features, encoded by the
dictionary $\Dbf$, in all the modalities. The second regularizer
in~\eqref{Eq:CostFunction} corresponds to the isotropic TV
penalty~\cite{Rudin1992}
\begin{equation}
 \Rcal(\xbf) \defn \sum_{\ell = 1}^L \sum_{n = 1}^N \|[\Lbf \xbf_\ell]_n\|_2,
\end{equation}
where $\Lbf$ denotes the discrete gradient operator. Unsupervised
learning of dictionaries from $\ybf$ is complicated when the imaging
problem is ill-posed. The goal of including the TV regularizer is to
assist this learning. In practice, we observed significant improvement
in quality when TV was included, both during learning and
reconstruction. Finally, the positive constants $\rho$, $\lambda$, and
$\tau$ are parameters controlling the tradeoff between the data
fidelity and regularization. \\
 
The joint optimization program in~\eqref{Eq:SparseCoding} is a convex
problem.  To solve it, we use the monotonic variant of FISTA
\cite{Beck2009}. In particular, we split $\Ccal( \xbf,\alphabm; \Dbf |
\Phibf, \ybf )$ into a smooth quadratic term
%
\begin{align}
 \tfrac{1}{2} \|  \ybf - \Phibf \xbf  \|_2^2  + \tfrac{\rho}{2} \|  \xbf - \Dbf \alphabm   \|_2^2 
\end{align}
and a non-smooth term that is separable in $\xbf$ and $\alphabm$
\begin{align}
    \la \norm{\alphabm }_{2,1} + \tau \Rcal(\xbf),
\end{align}
The proximal operator associated with $\la \norm{\bal }_{2,1}$ is equal to
\begin{align}
    \big[ \prox_{\la \norm{\cdot}_{2,1}}(\alphabm) \big]_{\cdot kn} =  \Big(\norm{ \alphabm_{\cdot kn} }_2 - \la  \Big)_+ \frac{ \alphabm_{\cdot kn} }{\norm{ \alphabm_{\cdot kn} }_2},
\end{align}
where the operator $(\cdot)_+$ extracts the positive part of its argument. While the proximal of TV does not have a closed-form solution, it can be efficiently implemented~\cite{Beck2009}.
\subsection{Learning Algorithm}
Suppose the input data is streamed so that at every time step $t \in
\NN$ we get a pair $(\ybf^{\iter{t}}, \Phibf^{\iter{t}})$. The
learning procedure attempts to minimize~\eqref{Eq:CostFunction} for
all $t$, jointly for $\xbf, \alphabm$ and $\Dbf$. Specifically, let
\allowbreak $\Jj^t (\Dbf) \defn \min_{\xbf, \bal} \left\{\Ccal(\xbf,
\alphabm; \Dbf \,|\, \ybf^{\iter{t}},\Phibf^{\iter{t}})\right\}$, then
this amounts to solving
\vspace{-1mm}
\begin{align}
\label{eq:dl-opt}
\min_{\Dbf \in \Dfrak} \left\{ \EE \left[ \Jj^t (\Dbf) \right] \right\},
\vspace{-1mm}
\end{align}
with respect to $\Dbf$, where the expectation is taken over $t$.
%
%
%
Note that, to compensate for scaling ambiguities, we restrict the optimization of $\Dbf$ to a closed convex set $\Dfrak$. Specifically, $\Dfrak$ is the set of convolutional dictionaries that have kernels in the $\ell_2$ ball, \ie $\norm{\dbf_{\ell k}}_2 \leq 1$.

The joint optimization program in \eqref{eq:dl-opt} is difficult to
solve directly. Thus, we use an alternating minimization procedure. In
particular, at iteration $t$, given the current dictionary
$\Dbf^{\iter{t-1}}$, and a new pair of data $(\ybf^{\iter{t}},
\Phibf^{\iter{t}})$, we first solve
\vspace{-1mm}
\begin{align}
\label{eq:sparse_coding}
(\xbf^{\iter{t}}, \alphabm^{\iter{t}}) \gets \uargmin{\xbf,\alphabm}  \left\{\Ccal(\xbf, \alphabm; \Dbf^{\iter{t-1}} \,|\, \ybf^{\iter{t}},\Phibf^{\iter{t}})\right\},
\vspace{-1mm}
\end{align}
using the method presented in Section~\ref{sec:problem}.  Then,
following the principle of \cite{Mairal2010}, we use all the previous
iterates and chose $\Dbf$ to minimize a surrogate of $\EE \left[ \Jj^t
  (\Dbf) \right]$ given by
\vspace{-1mm}
\begin{align}
    \frac{1}{t} \sum_{i=1}^t \Ccal(\xbf^{\iter{i}}, \alphabm^{\iter{i}}; \Dbf \,|\, \ybf^{\iter{i}},\Phibf^{\iter{i}}). 
\vspace{-1mm}
\end{align}
This second step, performed using a block gradient descent on the kernels $\{\dbf_{\ell k} \}$, is described in Section~\ref{sec:dict_update}.
The complete learning algorithm is summarized in Algorithm~\ref{alg:OnlineCDL}.
\begin{algorithm}[t]
\caption{Online Convolutional Dictionary Learning}\label{alg:OnlineCDL}

\begin{algorithmic}[1]
\Procedure{OnlineCDL}{}\\
\algorithmicrequire{ Stream of data $t \mapsto (\ybf^{\iter{t}},\Phibf^{\iter{t}})$, initial dictionary $\Dbf^{\iter{0}}$.}
\State $\Cbf^{\iter{0}} \gets \bs{0}$; \quad $\bbf^{\iter{0}} \gets 0$;
\While{streaming data,}
\State Draw a pair $(\ybf^{\iter{t}},\Phibf^{\iter{t}})$;
\State Sparse coding step:
\State $(\xbf^{\iter{t}}, \alphabm^{\iter{t}}) \gets \uargmin{\xbf,\alphabm}  \left\{\Ccal(\xbf, \alphabm; \Dbf^{\iter{t-1}} \,|\, \ybf^{\iter{t}},\Phibf^{\iter{t}})\right\}$ ;
\State Update memory: \vspace{1mm}
\State $\bbf^{\iter{t}} \gets    (1-\tfrac{1}{t}) \bbf^{\iter{t-1}}  +  \tfrac{1}{t} \sum_{i =1}^t \Abf^{\closeiter{i} \tr} \xbf^{\iter{i}}$; \vspace{2mm}
\State $\Cbf^{\iter{t}} \gets (1-\tfrac{1}{t})  \Cbf^{\iter{t-1}} + \tfrac{1}{t}  \sum_{i =1}^t \Abf^{\closeiter{i} \tr} \Abf^{\closeiter{i}} $; \vspace{2mm}
\State Dictionary update \eqref{eq:bloc_gradient}, \eqref{eq:projection} initialized with $\Dbf^{\iter{t-1}}$:\vspace{1mm}
\State $\Dbf^{\iter{t}} \gets \uargmin{\Dbf \in \Dfrak}   \tfrac{ 1 }{2t } \sum_{i = 1}^t \|  \xbf^{\iter{i}} - \Dbf \bal^{\iter{i}}   \|_2^2 $;
\EndWhile
\EndProcedure
\end{algorithmic}

\end{algorithm}

\subsection{Dictionary update}
\label{sec:dict_update}
Keeping $\xbf^{\iter{i}}$ and $\bal^{\iter{i}}$ fixed, the only term
in $\Ccal$ that depends on $\Dbf$ is the quadratic coupling penalty
${\tfrac{\rho}{2} \| \xbf^{\iter{i}} - \Dbf \bal^{\iter{i}}
  \|_2^2}$. Therefore, we can equivalently minimize
$
    \tfrac{1}{2t} \sum_{i=1}^t  \|  \xbf_{\ell}^{\iter{i}}  - \Dbf_{\ell} \alphabm_{\ell}^{\iter{i}}   \|_2^2,
$
for each modality $\ell$. Since everything is separable in $\ell$, in
the remainder we drop the subscript for notational clarity. Note that,
since the convolution operation is commutative and the
$\bal^{\iter{i}}$ are fixed, we can rewrite
\begin{flalign}
    \Dbf \bal^{\iter{i}} = \sum_{k=1}^{K} \dbf_{k} \ast \bal_{k}^{\iter{i}} =\sum_{k=1}^{K} \bal_{k}^{\iter{i}} \ast  \dbf_{k} = \Abf^{\closeiter{i}} \dbf,
\end{flalign}
where ${\Abf^{\closeiter{i}}} \defn (\Abf_{1}^{\closeiter{i} },\dots,\Abf_{K}^{\closeiter{i}}) \in \mathbb{R}^{N \times KP}$ is the sum-of-convolutions linear operator and $\dbf \defn \vert(\dbf_1,\dots, \dbf_K) $.
In order to minimize 
$
    \Gg^{\iter{t}}(\dbf) \defn \frac{1}{2t} \sum_{i=1}^t \|  \xbf^{\iter{i}}  - \Abf^{\closeiter{i}} \dbf  \|_2^2,
$
subject to $\norm{\dbf_{k}}_2 \leq 1$, as in \cite{Mairal2010}, we apply
a projected block-coordinate gradient descent. The algorithm starts
for $s=0$ with $\dbf^{\iter{t,0}} \gets \Dbf^{\iter{t-1}} $ and
iteratively applies the following two steps for all $k\in [1,\dots,K]$,
\begin{align}
   \label{eq:bloc_gradient}
    \tilde{\dbf}_{k}^{\iter{t,s}} &\gets  \dbf_{k}^{\iter{t,s-1}} - \frac{1}{L_k^{\iter{t}}} \nabla_{\dbf_k} \Gg^{\iter{t}}  \left( \vert(\dots, \tilde{\dbf}_{k-1}^{\iter{t,s}}, {\dbf}_{k}^{\iter{t,s-1}}, \dots ) \right) \\
   \label{eq:projection}
    \dbf_{k}^{\iter{t,s}} &\gets \tilde{\dbf}_{k}^{\iter{t,s}} / \max \{ \| \tilde{\dbf}_{k}^{\iter{t,s}}\|, 1 \}
\end{align}
%
until convergence or until a maximum number of iterations is reached. Note that $\nabla_{\dbf_k}$ denotes the partial gradient
\begin{align}
    \nabla_{\dbf_k} \Gg^{\iter{t}}(\dbf) =  \frac{1}{t} \sum_{i=1}^t  \Abf_{k}^{\closeiter{i} \tr} (\Abf^{\closeiter{i}}  \dbf - \xbf^{\iter{i}} ) ,
\end{align}
and $L_k^{\iter{t}} $ is the Lipschitz constant of $\nabla_{\dbf_k} \Gg^{\iter{t}}(\dbf)$. 
Importantly, we can take advantage of all the previous iterates to compute this gradient. Indeed, we can write it as 
\begin{align}
    \nabla_{\dbf_k} \Gg^{\iter{t}}(\dbf) =  \Cbf_{k}^{\iter{t}} \dbf - \bbf_{k}^{\iter{t}},
\end{align}
where the \emph{memory vector} and the symmetric \emph{memory matrix},
\begin{align}
    \bbf^{\iter{t}} =   \vert(
    \bbf_{1}^{\iter{t}} ,
    \dots,
     \bbf_{K}^{\iter{t}}) \defn \frac{1}{t} \sum_{i=1}^t \ \Abf^{\closeiter{i} \tr}\xbf^{\iter{i}},\\
    \Cbf^{\iter{t}} = \vert(
    \Cbf_{1}^{\iter{t}}
    ,\dots,
     \Cbf_{K}^{\iter{t}})
     \defn \frac{1}{t} \sum_{i=1}^t  \Abf^{\closeiter{i} \tr} {\Abf^{\closeiter{i}}},
\end{align}
%
with $\Cbf_k^{\iter{t}} \defn (\Cbf_{k,1}^{\iter{t}} ,\dots, \Cbf_{k,K}^{\iter{t}}) = ( \Abf_k^{\closeiter{i} \tr} \Abf_1^{\closeiter{i}},\dots,  \Abf_k^{\closeiter{i} \tr} \Abf_K^{\closeiter{i}}) $, are computed recursively from the previous iterates as
\begin{align}
\label{Eq:VecUpdate}
    \bbf^{\iter{t}} & \leftarrow \frac{t-1}{t}   \bbf^{\iter{t-1}}  + \frac{1}{t} \ \Abf^{\closeiter{t} \tr} \xbf^{\iter{t}},\\
\label{Eq:MatUpdate}
    \Cbf^{\iter{t}} & \leftarrow \frac{t-1}{t}  \Cbf^{\iter{t-1}} + \frac{1}{t}\ \Abf^{\closeiter{t} \tr} \Abf^{\closeiter{t}} .
\end{align}
Note that the aforementioned Lipschitz constant is $L_k^{\iter{t}}  = \norm{\Cbf^{\iter{t}}_{kk}}_2$.

\subsection{Implementation details}
\noindent\emph{Convolutional implementation.}\  A naive implementation of the online convolutional learning algorithm would require to store $\Cbf^{\iter{t}}$, which is a dense symmetric $KP \times KP$ matrix.
By definition, $\Abf_{k}^{\closeiter{i}} \in \RR^{N \times P}$ is a
convolution operator whose columns and rows are restricted to match
the size of its input $\dbf_k$ and output $\xbf$,
respectively. Similarly, $\Abf_{k}^{\closeiter{i} \tr} \in \RR^{P
  \times N} $ is a restricted convolution operator whose kernel
$\check{\bal}_{k}^{\iter{i}}$ is the flipped version of
$\bal_{k}^{\iter{i}}$. Therefore, if the size of $\bal_{k}^{\iter{i}}$
is chosen such that it implements a full convolution, then the operator $\Abf_{k}^{\closeiter{i} \tr} \Abf_{k'}^{\closeiter{i}}$ corresponds to a restricted convolution with a kernel $\check{\bal}_{k}^{\iter{i}} \ast \bal_{k'}^{\iter{i}}$. Importantly, the restrictions to a $P \times P$ operator imply that only part of the kernel, of size proportional to $ P$, is actually used. We denote this effective part by $\Sbf_P ( \check{\bal}_{k}^{\iter{i}} \ast \bal_{k'}^{\iter{i}})$ where $\Sbf_P$ is the selection operator. This implies that $\Cbf^{\iter{t}}_{kk'}$ is itself a convolution operator with kernel
\vspace{-1mm}
\begin{align}
\cbf_{kk'}^{\iter{t}} \defn \frac{1}{t} \sum_{i=1}^t \Sbf_P (\check{\bal}_{k}^{\iter{i}} \ast \bal_{k'}^{\iter{i}}) .
\end{align}
Therefore, an efficient implementation of $\Cbf^{\iter{t}}$ only requires to store or convolve with those $K^2$ kernels.
Note that this computational trick is a big argument in favor of decoupling $\frac{1}{2}\|\ybf - \Phibf \xbf\|_{2}^2$ from $\frac{1}{2}\|\xbf - \Dbf\alphabm\|_{2}^2$. By contrast, using the same technique to minimize
$
    \frac{1}{2} {\|  \ybf - \Phibf \Dbf \bal   \|_2^2}
$
instead, requires the explicit storage of a dense symmetric $KP \times KP$ matrix.
When $\Phibf$ is a mask and the dictionary is not convolutional, Mensch \emph{et al.}~\cite{Mensch2016} adopt a different strategy which consists in approximating the surrogate function. 

\medskip

\noindent\emph{Data centering.}\ Patch-based dictionary learning is
known to be more effective when the input data is centered to have
zero mean. While this is not required, it is common to pre-process the
data by removing the means of the training
patches~\cite{Mairal.etal2014}. Accordingly, we first estimate a local
mean, \ie, a low-pass component, $\xbf^{\lo} $ of the data
$\xbf$. The remaining component is the high-pass image $\xbf^{\hi}
\defn \xbf- \xbf^{\lo} $. We aim to learn the sparse synthesis model
$\xbf^{\hi} \approx \Dbf \bal$. To do so, we adapt the method above,
replacing the coupling term in \eqref{Eq:CostFunction} by
$\frac{\rho}{2}\|\xbf - \xbf^{\lo} -
\Dbf\alphabm\|_{2}^2$. Specifically, we use a mask-aware low-pass
filter to estimate $\xbf^{\lo}$ from $(\ybf,\Phibf)$ where $\Phibf$ is
a mask operator. Let $\Lcal(\cdot)$ be a low-pass filter and $\phibm
\in \{0,1\}^{N}$ the mask (\ie the diagonal of $\Phibf$). Then, for
every pixel $[\Lcal( \phibm )]_n > 0$, we compute
$[\xbf^{\lo}]_n = \frac{[\Lcal( \ybf )]_n} {[\Lcal( \phibm )]_n}$.
We use the nearest neighbor interpolation to fill the remaining pixels where $[\Lcal( \phibm )]_n = 0$.

\medskip

\noindent\emph{Mini-batch extension.}\ Similarly to~\cite{Mairal2010}, we enhance the algorithm by performing the sparse coding step on a few samples between every dictionary update. This is particularly appealing when we can process several samples in parallel.
It is worth noting that there is a trade off between the number of
input samples per mini-batch and their size. A big input sample
contains a lot of redundant information and leads to a slower coding
step. Conversely, a mini-batch of a few small but diverse samples is
faster and can mitigate the effect of a single iteration biasing
towards a specific scene.

\medskip

\noindent\emph{Forgetting factor.}\ In the first few iterations of the
algorithm, the initial dictionary may not be informative for effective
sparse representation. Thus, the corresponding coefficient maps
$\bal^{\iter{t}}$ are inaccurate, compared to coefficient maps
computed with later iterates. Consequently, we also introduce a
forgetting factor $\gamma \geq 0$, which allocates more weight to
newer samples than to older ones. In practice, during the update of
the memory vector \eqref{Eq:VecUpdate} and matrix
\eqref{Eq:MatUpdate}, we weigh the old and new ones respectively by
$\theta^{\iter{t}} \defn (1-\tfrac{1}{t})^{1+\gamma}$ and $1-
\theta^{\iter{t}}$.



\section{Experimental Evaluation}

\begin{figure}[t]
\centering
\includegraphics[width = 0.47\textwidth]{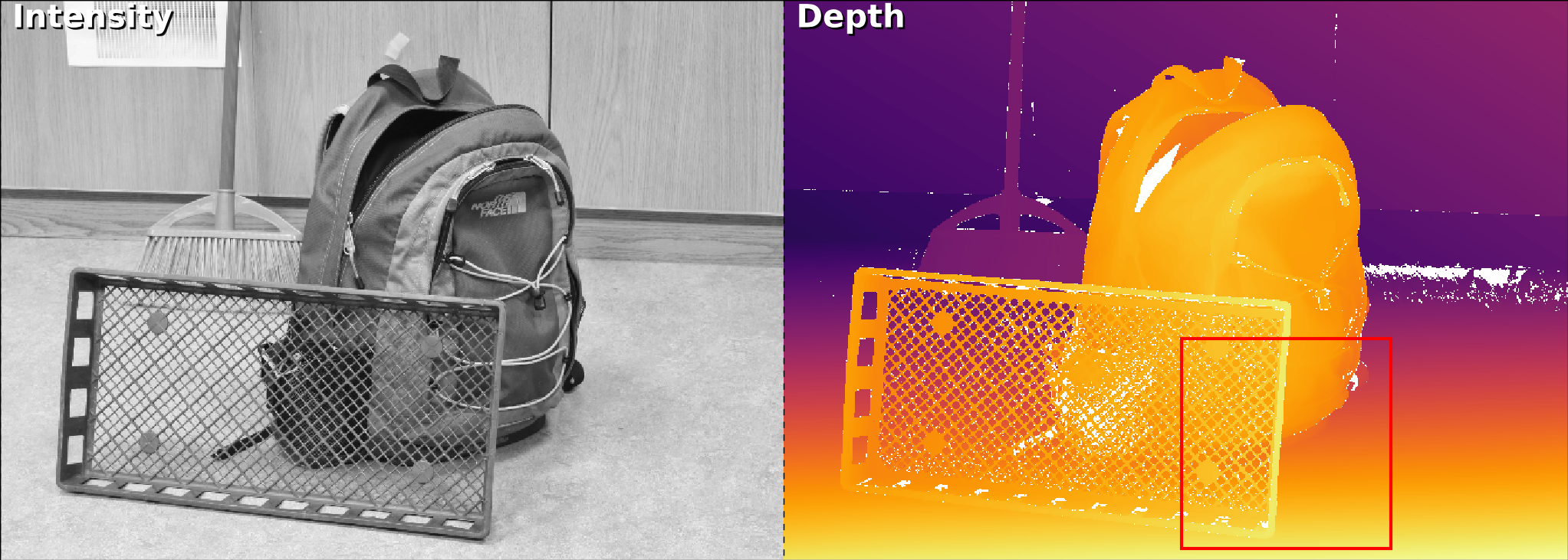}\\
\includegraphics[height = 0.28\textwidth]{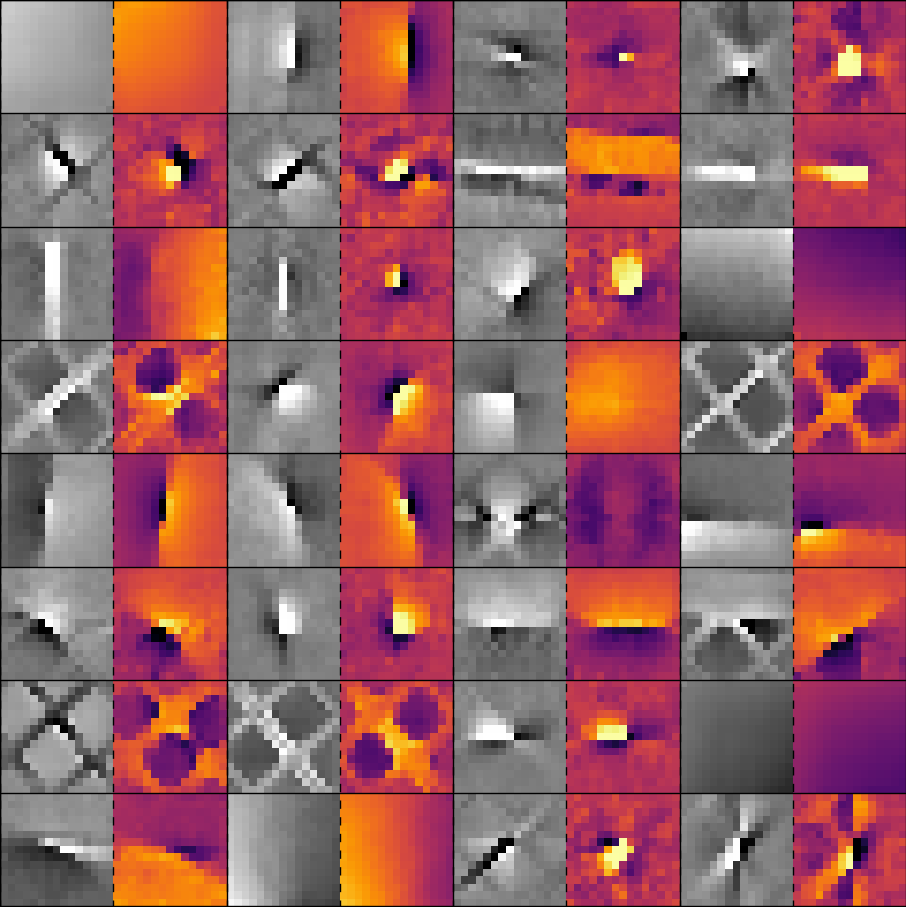}
\includegraphics[height = 0.28\textwidth]{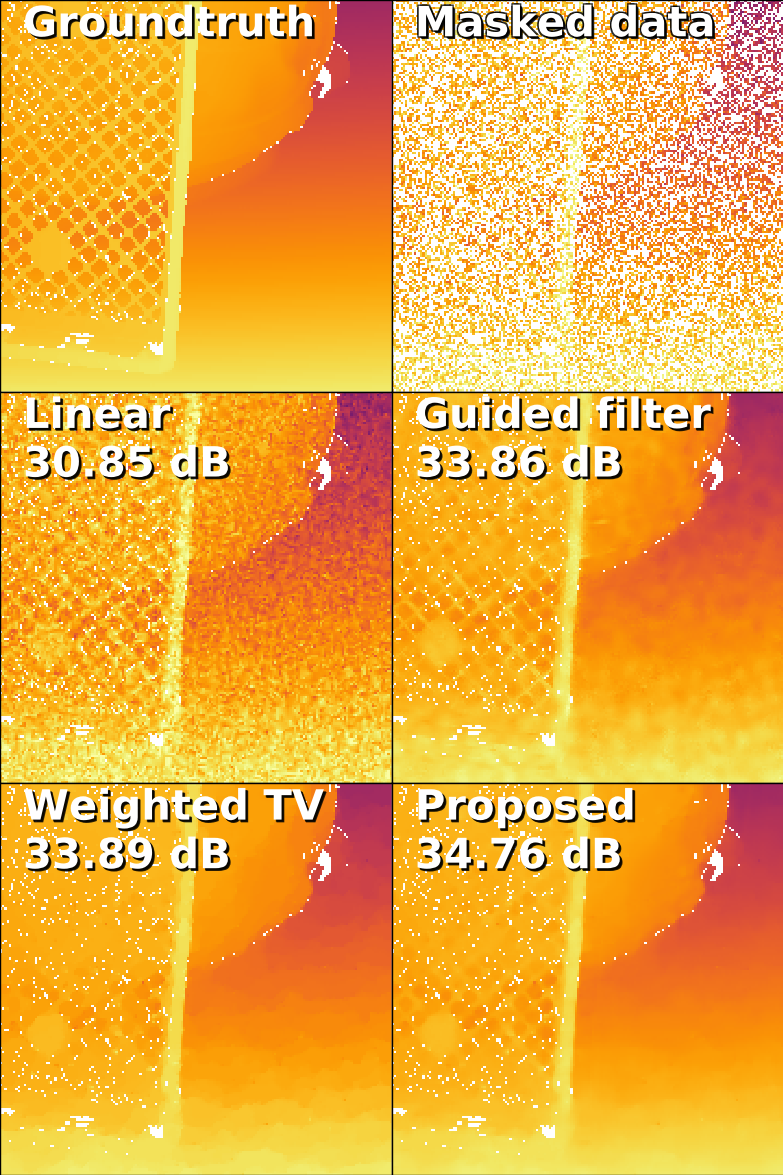}\vspace{-3mm}
\caption{The top-left and top-right images are  the intensity and depth modalities from the \emph{Backpack} image. White pixels correspond to missing pixels due, for example, to occlusions. Bottom-left image shows the trained dictionary with each pair of corresponding intensity-depth kernels grouped. Bottom-right shows the reconstruction of the region highlighted in red for $2 \times$ subsampling.}
\vspace{-2mm}
\label{fig:backpack}
\end{figure}
\begin{figure}[t]
\centering
\includegraphics[width = 0.45\textwidth]{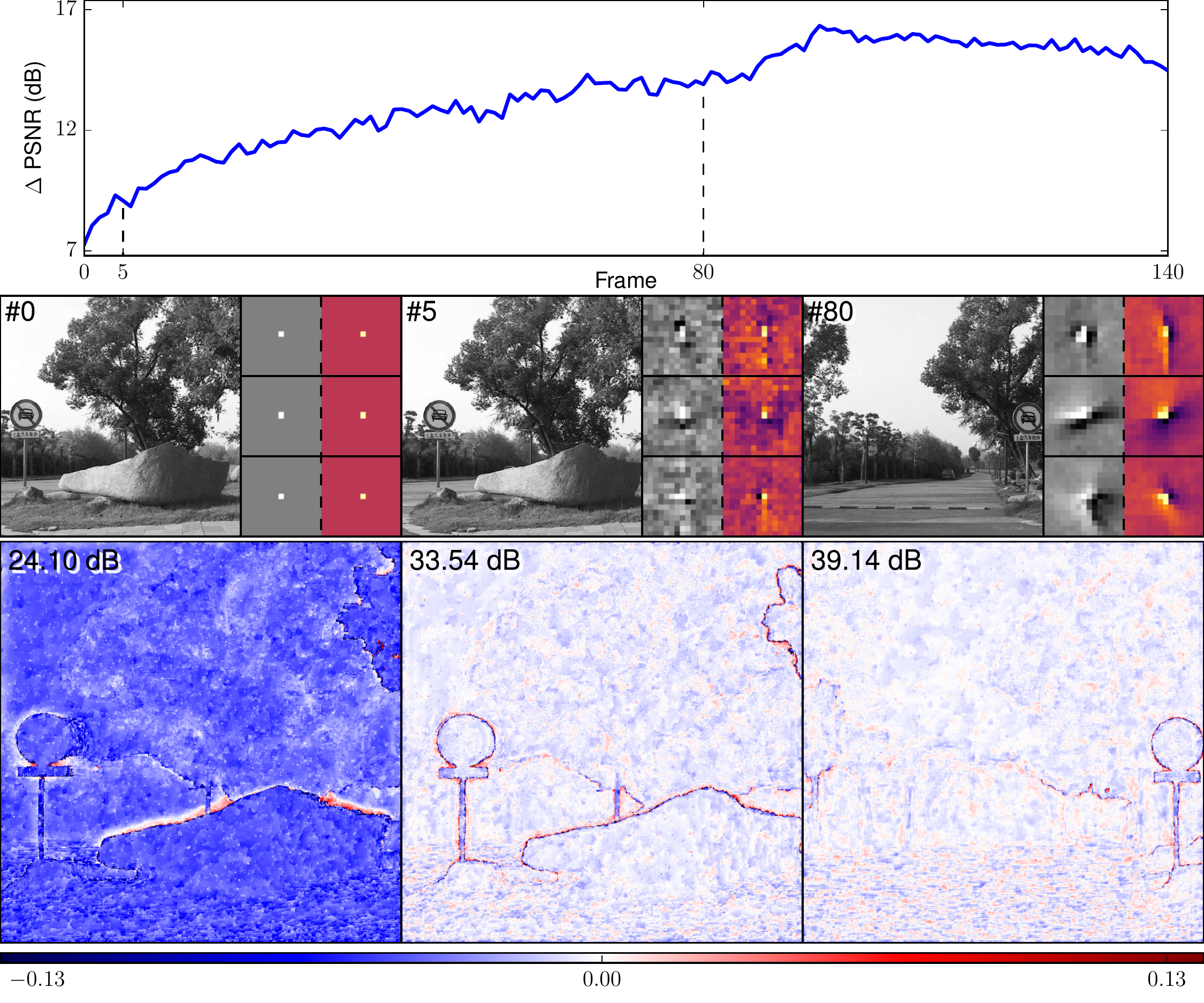}\vspace{-4mm}
\caption{Top row plots the evolution of the improvement in dB using the learned dictionary instead of deltas. Middle row shows the intensity frames 0, 5 and 80 next to three kernels from the corresponding dictionary. Bottom row shows the residual of the reconstructed depth. }
\vspace{-3mm}
\label{fig:psnr-iteration}
\end{figure}

To evaluate our multimodal imaging method, we focus on joint
intensity-depth reconstruction. We consider two modalities ($L=2$),
where $\Phibf_1 = \Id $ is the sensing matrix associated to an
intensity image and $\Phibf_2 = \diag(\phibm) $ is a random mask
selecting a fraction of the depth map pixels. To generate $\ybf_\ell$,
Gaussian noise of $30 $dB PSNR is added to $\Phibf_\ell \xbf_\ell$ on
both modalities. Our quality metric is the prediction PSNR over the
missing pixels of $\xbf_2$. Since the reconstruction method
\eqref{Eq:SparseCoding} optimizes over two distinct variables
($\xbfhat$,$\alphabmhat$), two options are available for the
prediction: either using $\xbfhat$ or using $\Dbf \alphabmhat +
\xbf^{\lo}$. In our experiments the second solution provides better
performance.
We rely only on subsampled data for both training and reconstruction,
and set the number of convolutional kernels to $K = 32$ and size $P =
15\times 15$.

\subsubsection*{Comparison with other methods}

\begin{table}[t]
\vspace{0mm}
  \caption{Average PSNR for  various subsampling rates on 23 images from the Middlebury 2014 dataset.}
  \label{table_comparison}
  \centering
\begin{tabular}{|l|c|c|c|}
  \hline
  \textbf{Method} & $\mathbf{2\times}$ & $\mathbf{3\times}$ & $\mathbf{4\times}$ \\ \hline
  Linear & $30.72$ dB       & $30.39$ dB & $29.97$ dB\\
  Guided Filter &  $34.18$ dB        & $33.46$ dB & $32.76$ dB  \\
  Weighted TV & $36.85$ dB        & $35.58$ dB & $34.77$ dB \\
  Proposed & $\mathbf{37.13}$ \textbf{dB}  & $\mathbf{35.73}$ \textbf{dB} & $\mathbf{34.88}$ \textbf{dB}  \\
  \hline
  \end{tabular}
  \vspace{-4mm}
\end{table}

In this experiment, we first train a global
dictionary using 160 mini-batches of 8 randomly selected patches of
size $45 \times 46$ from the Middlebury
dataset~\cite{Scharstein2014}. Then, to process each specific frame,
the dictionary is specialized with 120 mini-batches of 8 patches
sampled from that frame. Finally, the full ($480 \times 672$) frame is
reconstructed by solving~\eqref{Eq:SparseCoding} with the specialized
dictionary. Note that, in contrast to patch-based dictionary learning,
the input patches are not necessarily of the same size as the
dictionary kernels. In fact, the full frame could be used to train the
convolutional dictionary. However, using smaller patches instead,
helps to accelerate the training.

Table~\ref{table_comparison} compares the average performance of our
reconstruction procedure with three alternative approaches, listed in increasing complexity: linear interpolation, guided filtering~\cite{He.etal2013}, and weighted TV~\cite{Castorena2016a}. Note that, similarly to our method, the guided filter and the weighted TV both use intensity information as a guide for depth estimation. The parameters were hand-tuned using heuristics for every method in order to achieve the best average performance.

Figure~\ref{fig:backpack} shows the specialized dictionary and the reconstruction results for the \emph{Backpack} image. One can clearly recognize the typical image features manifested in the learned kernels. Some kernels present sharp edges or corners. Others show more elaborate gridded features. Most paired kernels have striking similarities between depth and intensity in terms of shape, orientation, and alignment. These results highlight the ability of our method to learn multimodal convolutional dictionaries directly from noisy and subsampled data.

\subsubsection*{Online training on a video}

In order to demonstrate the \emph{online} capability of our learning
algorithm, we use the \emph{Road} intensity-depth video sequence from
\cite{Zhang2009}. We use a mask with $2 \times$ subsampling and add a
$30$ dB Gaussian noise. We start with a dictionary filled with Dirac
deltas. On each $512 \times 512$ frame of the video, we extract 8
randomly chosen $50\times 50$ patches and perform one mini-batch
iteration of the learning algorithm. Then, we reconstruct the full
frame using the current dictionary and compare with the performances
obtained with the initial dictionary of
deltas. Figure~\ref{fig:psnr-iteration} presents the evolution, as the
video is streamed, of the PSNR improvement. As evident, the energy of
the residual decreases as the dictionary improves, especially near
object edges.  Note that a temporary drop of quality might be observed
when an unexpected feature appears in the scene. This decrease is then
compensated when the dictionary adapts to the new feature.

%


\bibliographystyle{IEEEbib}

\end{document}